\documentclass{article}

\usepackage[preprint, nonatbib]{neurips_2024}

\usepackage[utf8]{inputenc} 
\usepackage[T1]{fontenc}    
\usepackage{hyperref}       
\usepackage{url}            
\usepackage{booktabs}       
\usepackage{amsfonts}       
\usepackage{nicefrac}       
\usepackage{microtype}      
\usepackage{xcolor}         
\usepackage{amssymb}
\usepackage{mathtools}
\usepackage{amsthm}

\usepackage{algorithmic}
\usepackage{algorithm}

\usepackage{makecell}
\usepackage{graphicx}

\title{Adaptive Reasoning and Acting in Medical Language Agents}

\author{%
  Abhishek Dutta \\
  Department of Electrical and Computer Engineering\\
  University of Connecticut\\
  Storrs CT 06269, USA \\
  \And
  Yen-Che Hsiao \\
  Department of Electrical and Computer Engineering\\
  University of Connecticut\\
  Storrs CT 06269, USA \\
  \texttt{yen-che.hsiao@uconn.edu} \\
}

\begin{document}

\maketitle

\begin{abstract}
  This paper presents an innovative large language model (LLM) agent framework for enhancing diagnostic accuracy in simulated clinical environments using the AgentClinic benchmark. The proposed automatic correction enables doctor agents to iteratively refine their reasoning and actions following incorrect diagnoses, fostering improved decision-making over time. Experiments show that the implementation of the adaptive LLM-based doctor agents achieve correct diagnoses through dynamic interactions with simulated patients. The evaluations highlight the capacity of autonomous agents to adapt and improve in complex medical scenarios. Future enhancements will focus on refining the algorithm and expanding its applicability across a wider range of tasks and different large language models.
\end{abstract}

\section{Introduction}

Large language models (LLMs) have emerged as powerful statistical tools capable of predicting the next word, phrase, or even entire paragraphs based on the given input \cite{demszky2023using}. The effectiveness of these models can significantly depend on the prompts they receive \cite{arvidsson2023prompt}. One notable feature of LLMs is in-context learning, allowing them to grasp new tasks from a few examples provided within the prompt during inference \cite{minaee2024large}. This leads to the practice known as prompt engineering, which involves crafting and refining input prompts to elicit the desired responses from these models \cite{ekin2023prompt}.

The application of large language models (LLMs) in healthcare has demonstrated significant potential, with models achieving remarkable results on tasks such as the GPT-4 \cite{achiam2023gpt} achieves the average accuracy of around 83.15 from the United States Medical Licensing Examination (USMLE) self assessment dataset in \cite{nori2023capabilities}. However, in real-world clinical practice, diagnosis is a dynamic process involving continuous patient interaction, ordering of medical tests, and decision-making under uncertainty. Simulated clinical environments offer a valuable way to evaluate these models in more interactive, adaptive settings that reflect the realities of patient care.

In this paper, we leverage AgentClinic \cite{schmidgall2024agentclinic}, a multimodal benchmark designed to simulate clinical environments, to assess the performance of LLM agents in diagnosing patients through iterative doctor-patient dialogue, medical test interpretation, and bias management. AgentClinic \cite{schmidgall2024agentclinic} features four agents: the Doctor Agent, responsible for gathering information and making diagnoses; the Patient Agent, which simulates real-world patient interactions; the Measurement Agent, which provides test results; and the Moderator Agent, which evaluates the accuracy of the diagnosis. This setup allows for a detailed analysis of how LLM agents perform in sequential decision-making processes.

A key focus of this work is on handling cases where the doctor agent fails to provide an accurate diagnosis. We propose an automatic correction framework that enables the doctor agent to iteratively refine its reasoning after an incorrect diagnosis, ultimately arriving at the correct diagnosis through subsequent interactions. This framework introduces an adaptive feedback loop that adjusts the decision-making process of the doctor agent, allowing it to learn from its mistakes and correct itself over time.

Our contributions are as follows: Firstly, We introduce a robust adaptation mechanism for doctor agents that reason/act and observe, enabling them to improve diagnostic accuracy after initial failures. This system guides the doctor agent through a process of adaptive reasoning, helping it to correct earlier mistakes and reach a proper diagnosis. Secondly, we evaluate this framework in the AgentClinic \cite{schmidgall2024agentclinic} environment, demonstrating how it enhances the doctor agent’s ability to recover from incorrect diagnoses and improves overall diagnostic performance through adaptive learning.

Our work highlights the potential of autonomous agents in healthcare, showcasing how they can enhance diagnostic processes by enabling the doctor language agent to iteratively refine its reasoning and ultimately arrive at a correct diagnosis.

\section{Simulated clinical environment}

The AgentClinic benchmark \cite{schmidgall2024agentclinic} is a simulated clinical environment designed to evaluate the performance of AI models, particularly large language models (LLMs), in tasks that require real-time decision-making and patient interaction, mimicking the complexities of clinical settings. Unlike traditional static medical question-answering tests, this benchmark incorporates a more dynamic and interactive approach by simulating dialogues between patient and doctor agents, along with medical exams and tests, through multimodal agents.

In AgentClinic \cite{schmidgall2024agentclinic}, four main agents simulate the clinical environment: (1) Doctor Agent: The model being evaluated for its diagnostic abilities. This agent begins with minimal context about a patient’s condition and must interact with the patient agent to gather relevant information. It can ask a limited number of questions, request specific medical tests via the measurement agent, and ultimately provide a diagnosis. This setup simulates the process of sequential medical decision-making, requiring the doctor agent to operate under realistic clinical constraints, such as finite time and limited diagnostic resources. (2) Patient Agent: The patient agent holds information about symptoms, medical history, and lifestyle but does not know the final diagnosis. Its role is to provide responses that emulate real patient behavior during doctor-patient consultations. The patient agent can exhibit cognitive and implicit biases, affecting its interaction with the doctor agent. These biases emulate real-world patient biases, such as self-diagnosis based on internet research or distrust of the doctor based on implicit factors. (3) Measurement Agent: This agent simulates diagnostic tests, providing realistic medical readings based on the patient’s condition. For example, it can deliver results from an electrocardiogram, blood pressure readings, or imaging tests like X-rays. The doctor agent can request specific tests, and the measurement agent responds with results that match the patient’s simulated condition, contributing to the decision-making process. (4) Moderator Agent: This agent evaluates the doctor agent’s performance, determining whether the correct diagnosis has been made based on the information gathered during the interaction. The moderator ensures the dialogue is parsed correctly and compares the diagnosis with the actual medical condition to assess the accuracy of the doctor agent.

AgentClinic \cite{schmidgall2024agentclinic} also includes biases in the behavior of both patient and doctor agents, allowing researchers to study the impact of cognitive and implicit biases on medical decision-making. The benchmark introduces various patient types, with 107 patient agents having unique family histories, age groups, diseases, and lifestyle habits. 

\section{Autonomous Agent Architecture}

Let a simulated clinical environment be denoted as a function $f$ that maps a state $s\in\mathbb{V}$ and an action $a\in\mathbb{V}$ to an observation $o\in\mathbb{V}$, where $\mathbb{V}$ is a set of vocabulary. Let $\pi_\theta$ be an LLM agent over a pre-trained set of parameters $\theta$. Let $s_0$ be the initial state of the environment $f$, we aim to produce a sequence of actions $(a_0,a_1,a_2,\dots)$, where $a_i\in\mathbb{V}$ for $i\in\mathbb{Z}$, from a doctor LLM agent to change the state to a terminal state that indicates the patient is correctly diagnosed. 

\begin{figure}[!t]
\centering
\includegraphics[width=0.98\textwidth]{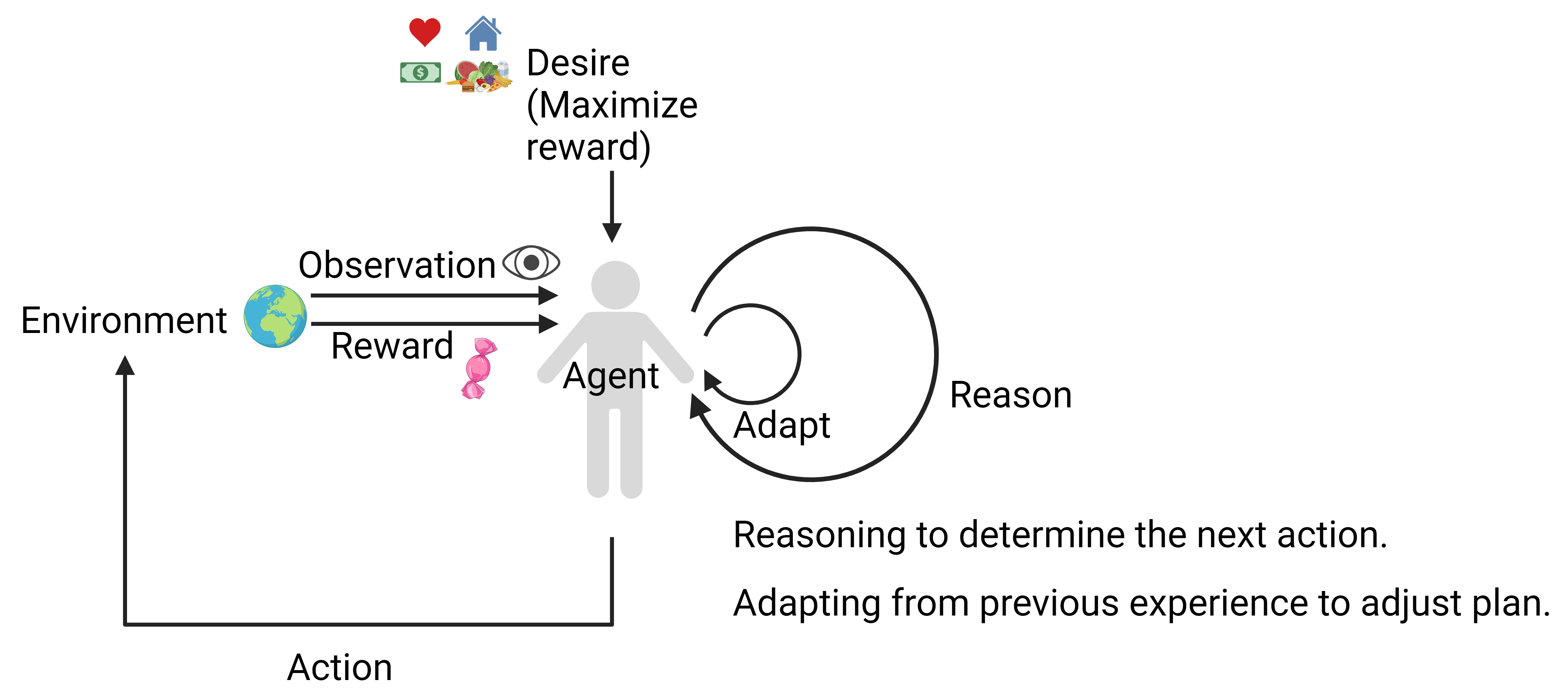}
\caption{An architecture towards autonomous agent. Created with BioRender.com.}
\label{fig:architecture}
\end{figure}

The architecture of the main idea of our work is shown in Figure \ref{fig:architecture}. A desire is provided to an agent to motivate it to solve a specific task in a given environment. The agent can perform an action to interact with the environment, causing the state of the environment to change. The agent then receives an observation that describes the status of the environment and a reward signal. The action may be proposed from two different processes: the reasoning process determines the next action based on the current progress; and the adaptation process summarizes previous progress to provide a better plan towards maximizing the reward.

We present a novel algorithm in Algorithm \ref{alg1}. Initially, we have the initial state $s_0$ which provides instructions, presents exemplars, and describes the environment and the goal for a specific task. $\pi_\theta$ is an LLM agent with a set of parameters $\theta$. $\tau=\{s_0,a_0,o_1,\dots\}$ is a sequence of the concatenation of state, action, and observation, where $s_k$, $a_k$, and $o_k$ are sequences of tokens representing the $k$-th state, action, and observation for $k\in\mathbb{Z}$, respectively. The return $R(\tau)$ is a string indicating whether the task is completed or not. $ep$ is a variable indicating the number of trials. The environment is reinitialized at each trial. 

Initially, for the doctor agent, we have the initial state $s_0$ which contains some context about what is known about the patient as well as a brief objective. $\pi_\theta$ is an LLM agent with a set of parameters $\theta$. $\tau=\{s_0,a_0,o_1,\dots\}$ is a sequence of the concatenation of state, action, and observation, where $s_k$,  $a_k$, and $o_k$ are sequences of tokens representing the $k$-th state, action, and observation for $k\in\mathbb{Z}$, respectively. The return $R(\tau)$ is a string indicating whether the task is completed or not. $ep$ is a variable indicating the number of trials. The environment is reinitialized at each trial.

At the first time step $k=0$, the action is then sampled from
\begin{equation}
a^1_0\sim\pi_{\theta}(a^1_0|s^1_0),
\end{equation}
where $a^1_0$ is a sequence of tokens which represents the first action in the first trial, $s^1_0$ is a sequence of tokens which represents the first state in the first trial, the subscript $0$ indicates the first time step, and the superscript $1$ indicates the first trail. The observation in the first trial, $o^1_1$, is a sequence of tokens obtained from the response of either a patient agent or a measurement agent. The observation can be represented by executing the action $a^1_0$ in the environment $f$ at state $s^1_0$ as
\begin{equation}
o^1_1=f(s^1_0, a^1_0).
\end{equation} 
A new state $s^1_1$ is formed by concatenating the action $a^1_0$ and the observation $o^1_1$ after state $s^1_0$ as
\begin{equation}
s^1_1=\{s^1_0,a^1_0,o^1_1\}.
\end{equation} 
If a maximum time step is reached or the doctor agent provides an incorrect diagnosis, the task fails and the return $R(\tau)$ is concatenated with self correction to form the initial state in the next trial $s^2_0$ as
\begin{equation}
s^2_0=\{R(\tau)\},
\end{equation} 
where $\tau=\{s_0,a_0,o_1,a_1,o_2,\dots o_{50}\}$. In the next trial, a sequence of tokens is generated from the LLM by 
\begin{equation}
t^2_0\sim\pi_{\theta}(t^2_0|s^2_0),
\end{equation} 
We call $t^{ep}_0$ at the $ep$-th trial for $ep>1$ as the adaptation from the ($ep-1$)-th trail and $t^{ep}_0$ indicates the correction of the ($ep-1$)-th failed trail to improve the next trail. In the next step, we propose to replace the initial state in the second trial with the initial state from the first trail to remove the dialogue from the previous trial such that the context length is reduced. We call this step compression. By performing compression, the first action in the second trail will only be conditioned on the initial state in the first trail $s^1_0$ and the adaptation from the first trail $t^2_0$ as 
\begin{equation}
a^2_0\sim\pi_{\theta}(a^2_0|s^2_0, t^2_0).
\end{equation} 

\begin{algorithm}[H]
\caption{Adaptive reasoning and acting}\label{alg:alg1}
\begin{algorithmic}
\STATE Initialize the world state $s_0$ as a text of exemplars and task, where each token $\in Vocab$.
\STATE Let $\pi_\theta$ be a LLM agent over a pre-trained set of parameters $\theta$.
\STATE Let a trajectory $\tau=\{s_0,a_0,o_1,\dots\}$ be a sequence of state, action, and observation.
\STATE Let $R(\tau)$ be the return for trajectory $\tau$.
\STATE Let $ep=1$.
\STATE While $R(\tau)\neq"OK"$ do
\STATE \hspace{0.5cm} Let $k=0$.
\STATE \hspace{0.5cm} While $k<50$ $||$ $R(\tau)="OK"$ do
\STATE \hspace{1cm} If $ep>1$ and $k=0$, then generate adaptation $t^{ep}_k\sim\pi_{\theta}(t^{ep}_k|s^{ep}_k)$.
\STATE \hspace{1cm} Compression step:
\STATE \hspace{1.5cm} If $k=0$, then $s^{ep}_0=s_0$.
\STATE \hspace{1cm} If $ep>1$, then generate action $a^{ep}_k\sim\pi_{\theta}(a^{ep}_k|s^{ep}_k, t^{ep}_0)$.
\STATE \hspace{1cm} If $ep=1$, then generate action $a^{ep}_k\sim\pi_{\theta}(a^{ep}_k|s^{ep}_k)$.
\STATE \hspace{1cm} Get observation $o^{ep}_{k+1}=f(s^{ep}_k, a^{ep}_k)$.
\STATE \hspace{1cm} Let $s^{ep}_{k+1}=\{s^{ep}_{k},a^{ep}_k,o^{ep}_{k+1}\}$.
\STATE \hspace{1cm} $k:=k+1$
\STATE \hspace{0.5cm} Concatenate $R(\tau)$ with "New plan: ".
\STATE \hspace{0.5cm} $s^{ep+1}_0=\{s^{ep}_k,R(\tau)\}$
\STATE \hspace{0.5cm} $ep:=ep+1$
\end{algorithmic}
\label{alg1}
\end{algorithm}

\section{Experimental results}

\begin{figure}[!t]
\centering
\includegraphics[width=0.98\textwidth]{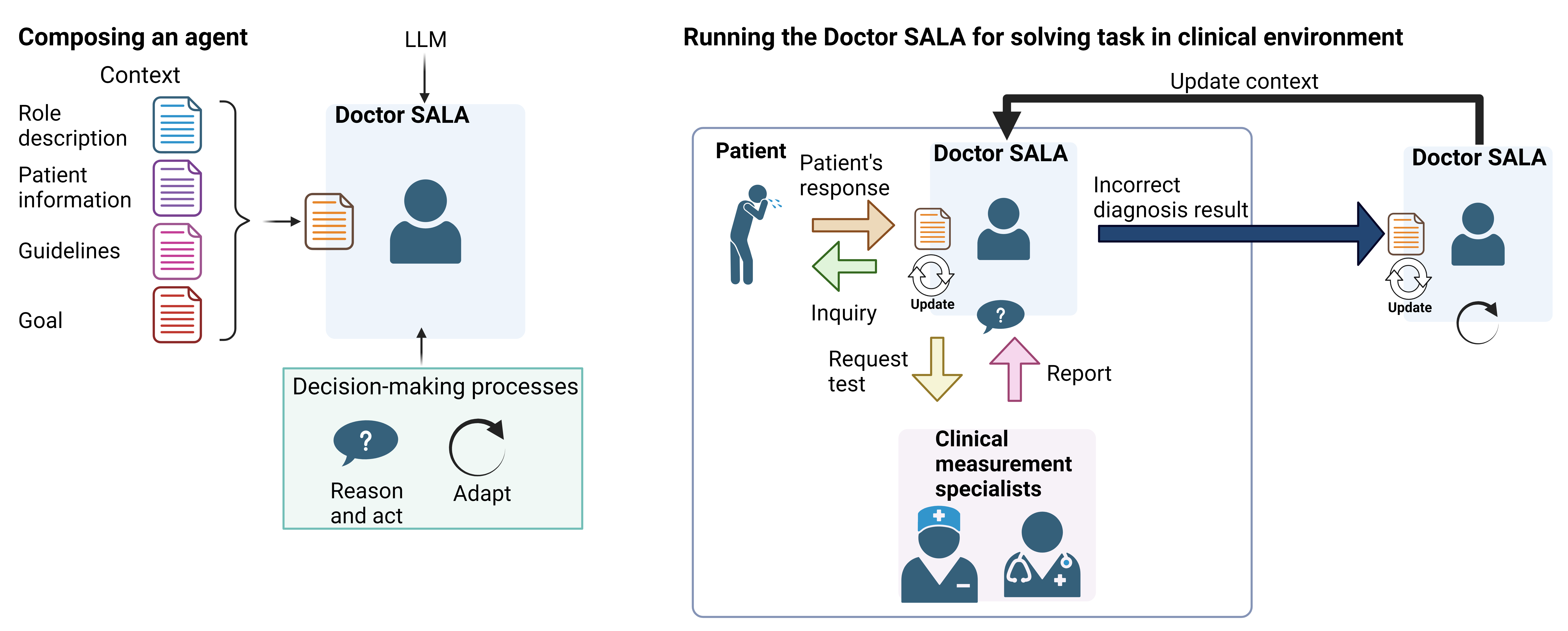}
\caption{Medical Self-Adaptive Language Agent. Created with BioRender.com.}
\label{fmedsala}
\end{figure}

We conducted experiments on the MedQA dataset from \cite{schmidgall2024agentclinic}, utilizing 15 scenarios with a maximum of 20 inferences, without bias or image requests, employing GPT-4 \cite{achiam2023gpt} as the patient, measurement, and moderator language agent. In the first experiment, we used both GPT-4 \cite{achiam2023gpt} and GPT-3.5 \cite{brown2020language} as the doctor language agent policy $\pi_{\theta}$ to compare the diagnostic results from these two different models. In each step of the sequence of play, the doctor agent, based on the given context state $s_0$, takes action $a_0$ which can be either to consult the patient agent or invoke the measurement agent, whose replies become the observation $o_1$. Now based on this added context, the doctor agent takes the next action $a_1$ and the cycle continues till the doctor makes the diagnosis or fails, which is the return $R(\tau)$. This medical self-adaptive language agent is presented in Figure \ref{fmedsala}.

\begin{figure}[!t]
\centering
\includegraphics[width=1\textwidth]{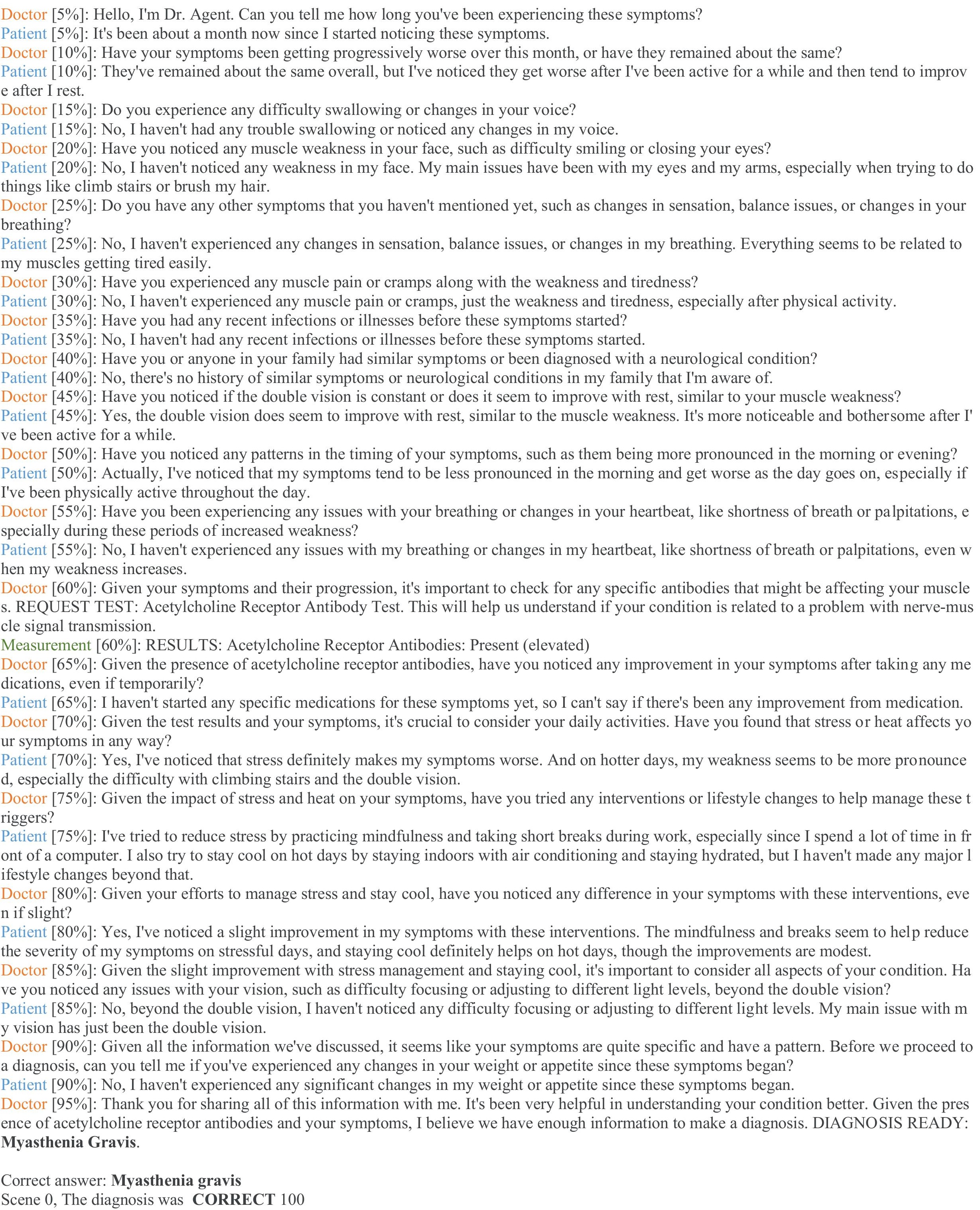}
\caption{The clinical consultation dialogue of the first case in the MedQA simulated clinical environment in \cite{schmidgall2024agentclinic} from a GPT-4 \cite{achiam2023gpt} doctor, patient, measurement, and moderator language agent. The doctor correctly diagnosed the patient with Myasthenia Gravis.}
\label{fig:GPT4Result1}
\end{figure}

\begin{figure}[!t]
\centering
\includegraphics[width=1\textwidth]{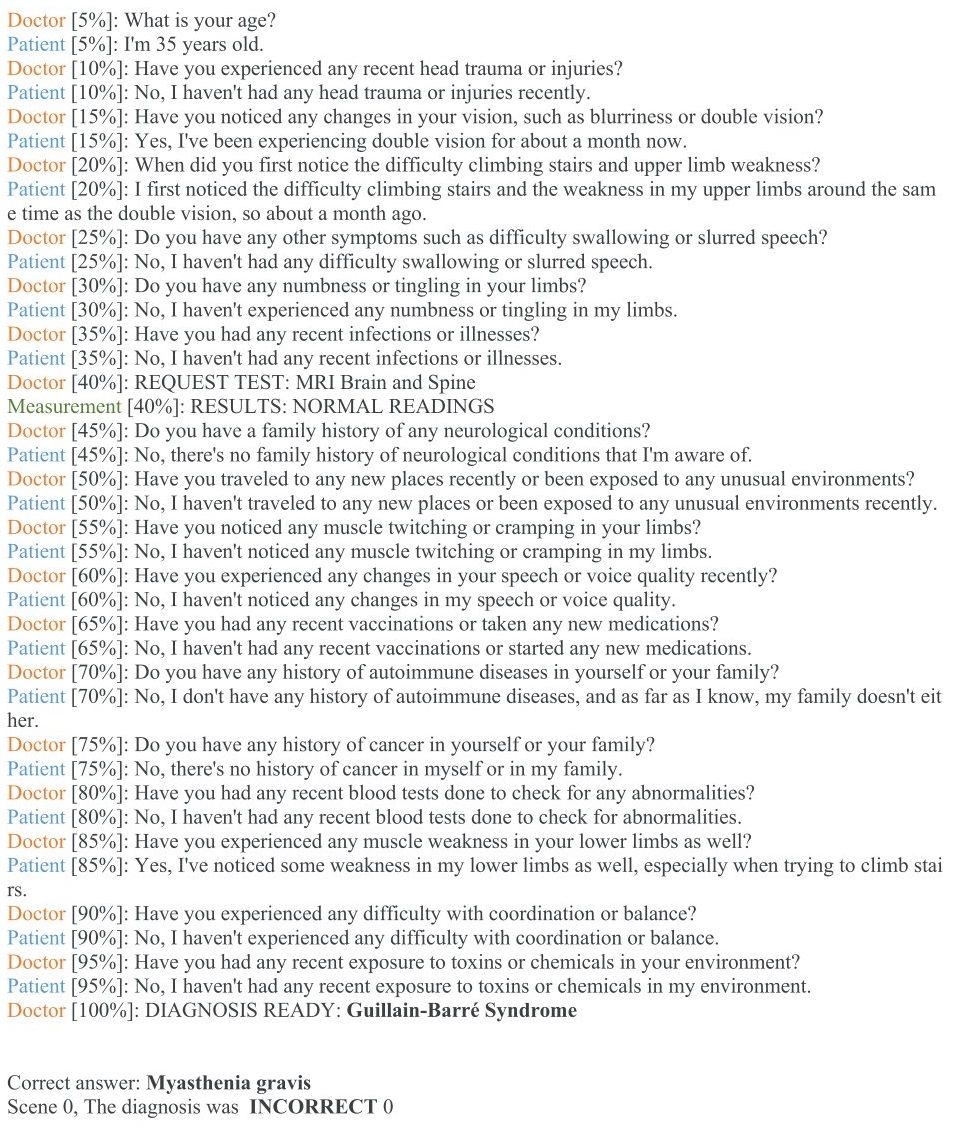}
\caption{The clinical consultation dialogue of the first case in the MedQA simulated clinical environment in \cite{schmidgall2024agentclinic} from a GPT-3.5 \cite{brown2020language} doctor language agent and GPT-4 patient, measurement, and moderator language agent. The doctor misdiagnosed the patient with Guillain-Barré Syndrome, but the patient actually has Myasthenia Gravis.}
\label{fig:GPT3_5Result1}
\end{figure}

The results are presented in Table~\ref{tab:Inference}. Now, for the first case in the MedQA simulated clinical environment, the GPT-4 \cite{achiam2023gpt} doctor $\pi_{gpt-4}$ comes with the right diagnosis as show in in the clinical consultation dialogue of Figure~\ref{fig:GPT4Result1}. However, the GPT-3.5 \cite{brown2020language} doctor $\pi_{gpt-3.5}$ in the same case fails to get to the correct diagnosis as shown in the clinical consultation dialogue of Figure~\ref{fig:GPT3_5Result1}.

\begin{table*}[!t]
\caption{Diagnosis result of different language models from the MedQA simulated clinical environment in \cite{schmidgall2024agentclinic} \label{tab:Inference}}
\centering
\begin{tabular}{|c|c||c|c|}
\hline
\makecell{Task\\number} & Correct answer & GPT-4 \cite{achiam2023gpt} & GPT-3.5 \cite{brown2020language}\\
\hline
1 & Myasthenia gravis & Correct & Guillain-Barré Syndrome \\
\hline
2 & \makecell{Progressive multifocal \\encephalopathy} & Correct & No answer \\
\hline
3 & Hirschsprung disease & Congestive Heart Failure & Correct \\
\hline
4 & \makecell{Diffuse large \\B-cell lymphoma} & Correct & Hodgkin's lymphoma \\
\hline
5 & \makecell{Acute interstitial \\nephritis} & Correct & Pyelonephritis \\
\hline
6 & Pes anserine bursitis & Patellar tendinitis & \makecell{Patellofemoral \\Pain Syndrome} \\
\hline
7 & Situational syncope & Correct & No answer \\
\hline
8 & \makecell{Congenital \\Rubella Infection} & CHARGE Syndrome & \makecell{Blue Rubber Bleb\\ Nevus Syndrome} \\
\hline
9 & Phyllodes tumor & Fibroadenoma of the breast & breast cancer \\
\hline
10 & Endometritis & Correct & Correct \\
\hline
11 & Hemorrhoids & Correct & Rectal Prolapse \\
\hline
12 & Complex partial seizure & Correct & Correct \\
\hline
13 & Posterior hip dislocation & Correct & Correct \\
\hline
14 & Hirschsprung’s disease & Correct & Correct \\
\hline
15 & \makecell{Rupture of the \\flexor digitorum \\profundus tendon \\at its point of insertion} & \makecell{Sprain of the \\distal interphalangeal joint \\of the right ring finger \\with possible ligament injury} & No answer \\
\hline
\end{tabular}
\end{table*}

\begin{figure}[!t]
\centering
\includegraphics[width=1\textwidth]{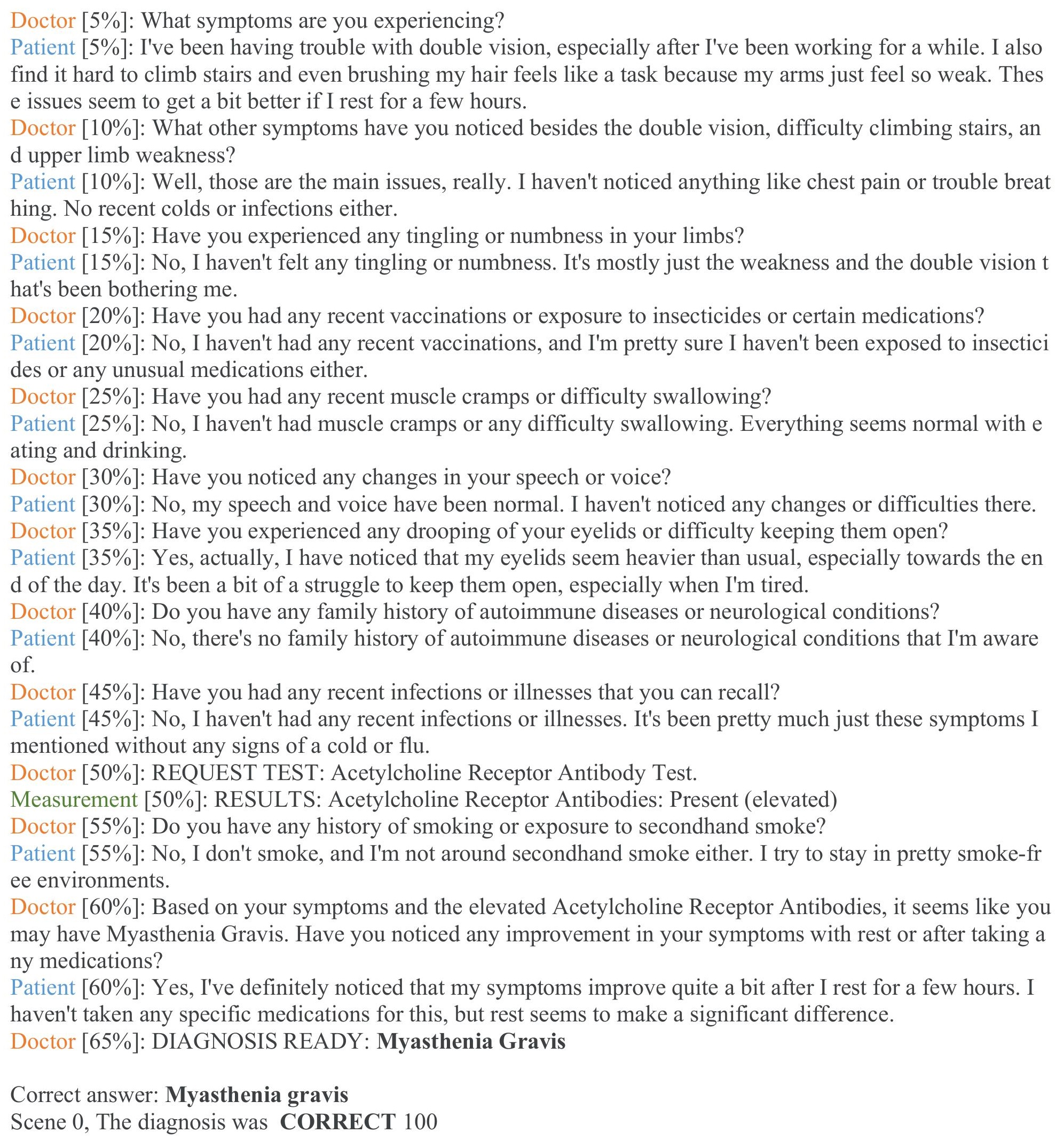}
\caption{The clinical consultation dialogue of the first case in the MedQA simulated clinical environment in \cite{schmidgall2024agentclinic} from a GPT-3.5 \cite{brown2020language} doctor language agent and GPT-4 patient, measurement, and moderator language agent using our proposed method. The doctor correctly diagnosed the patient with Myasthenia Gravis with 1 test and 12 questions, which is less than the doctor agent in Firgure~\ref{fig:GPT4Result1} with 1 test and 19 questions.}
\label{fig:GPT3_5_ManualRflx1_Result1_Crop}
\end{figure}

Therefore, per our algorithm, a correction or adaptation $t_0$ is added to the initial context $s_0$ as an exemplar and the cycle is repeated. As shown in Figure~\ref{fig:GPT3_5_ManualRflx1_Result1_Crop}, by adding the reflection, "If the patient has symptoms such as double vision, difficulty climbing stairs, and upper limb weakness, perform an Acetylcholine Receptor Antibody Test instead of an MRI of the brain and spine.", to the system prompt of the doctor agent, the $\pi_{gpt-3.5}$ doctor agent can correctly diagnose Myasthenia Gravis from the patient with 1 test and 12 questions, which is less than the $\pi_{gpt-4}$ doctor agent in Figure~\ref{fig:GPT4Result1} with 1 test and 19 questions.

\section{Conclusion}

In this paper, we have explored the capabilities of large language model (LLM) agents in a simulated clinical environment through the MedQA simulated clinical environment in AgentClinic \cite{schmidgall2024agentclinic}. By leveraging the power of in-context learning together with reason/act and observe, we introduced an automatic correction mechanism for doctor agents, enabling them to enhance their diagnostic accuracy after initial failures. Our experiments demonstrated that this framework can help the LLM doctor agent to achieve correct diagnoses over time, even in the face of complex patient interactions and decision-making scenarios.

The results from our evaluations highlight the significant potential of autonomous agents in healthcare settings, particularly in mimicking the dynamic nature of clinical practice. As we advance the field of AI in medicine, our findings underscore the importance of developing intelligent systems that can learn from experience and continuously improve their performance.

For future work, we aim to extend the framework’s applicability by incorporating a wider variety of tasks, such as differential diagnosis and treatment recommendations, to assess the versatility of the LLMs in dynamic clinical interactions. 

Furthermore, we intend to explore the performance of various large language models, comparing their capabilities in the AgentClinic framework. This comparative analysis will help identify the most effective models for specific diagnostic tasks and provide insights into their strengths and limitations in healthcare settings. By continuously improving our algorithms and expanding the tasks performed by the agents, we seek to develop more sophisticated autonomous systems that can significantly contribute to enhancing patient care and clinical decision-making.

\bibliographystyle{unsrt}
\bibliography{sample}






\end{document}